%% file: main.tex
\documentclass[letterpaper, 10pt, conference]{ieeeconf}  

\IEEEoverridecommandlockouts                              

\usepackage{graphics} 
\usepackage{epsfig} 
\usepackage{times} 
\usepackage{amsmath} 
\usepackage{amssymb}  
\usepackage[bookmarks=true]{hyperref}
\usepackage{xcolor,colortbl}
\usepackage{color}
\usepackage{siunitx}
\usepackage{multirow}
\usepackage{multicol}
\usepackage[skip=0.3\baselineskip]{caption}
\usepackage{booktabs}
\usepackage{pifont}
\usepackage{cite}
\usepackage{bm}
\usepackage{subcaption}
\usepackage{CJKutf8}
\usepackage{tikz}
\usepackage{float}
\usetikzlibrary{svg.path}
\usepackage{balance}

\usepackage{tabularx}
\newcolumntype{C}[1]{>{\centering\arraybackslash}p{#1}}

\usepackage{footnote}
\makesavenoteenv{tabular}
\makesavenoteenv{table*}

\usepackage[utf8]{inputenc}
\usepackage{makecell}
\usepackage{pgfplots}
\DeclareUnicodeCharacter{2212}{−}
\usepgfplotslibrary{groupplots,dateplot}
\usetikzlibrary{patterns,shapes.arrows}
\pgfplotsset{compat=newest}

\newcommand{\rebuttal}[1]{{\color{black}#1}}

\let\oldtwocolumn\twocolumn
\renewcommand\twocolumn[1][]{%
    \oldtwocolumn[{#1}{
    \begin{center}
    \begin{tikzpicture}
    \node[anchor=south west,inner sep=0] at (0,0) {
    \includegraphics[width=0.587\textwidth]{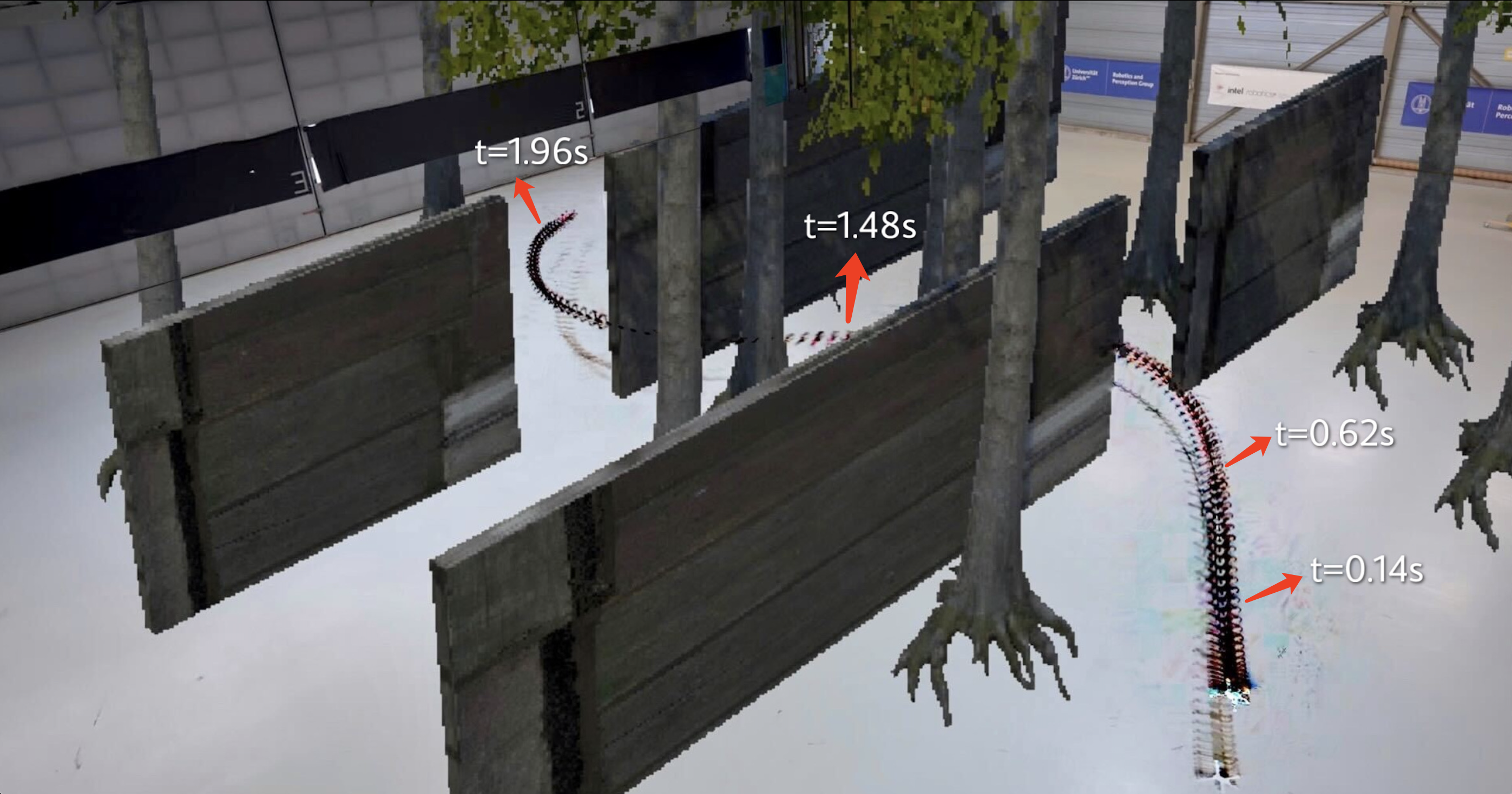}
    };
    \node[anchor=south west,inner sep=0] at (10.47,2.76) {
    \includegraphics[width=0.204\textwidth]{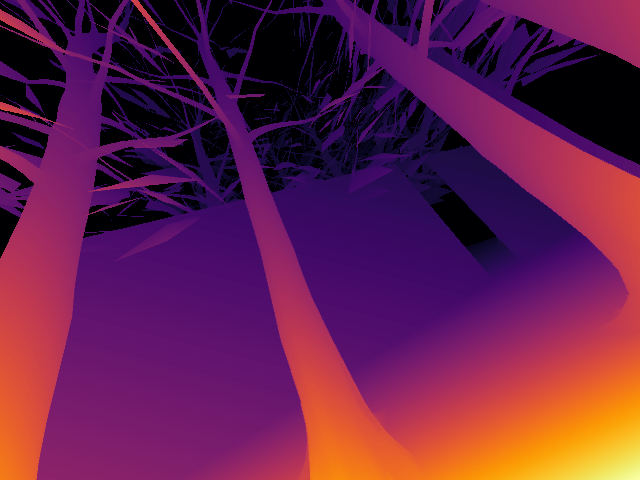}
    };
    \draw[anchor=south west] (10.47,2.76) node [text=white] {\small t~=~\SI{0.14}{\second}};

    \node[anchor=south west,inner sep=0] at (14.13,0) {
    \includegraphics[width=0.204\textwidth]{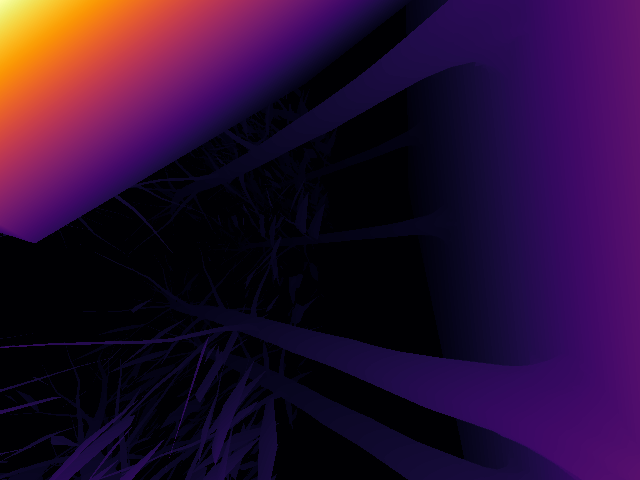}
    };
    \draw[anchor=south west] (14.13,0) node [text=white] {\small t~=~\SI{1.96}{\second}};

    \node[anchor=south west,inner sep=0] at (10.47,0) {
    \includegraphics[width=0.204\textwidth]{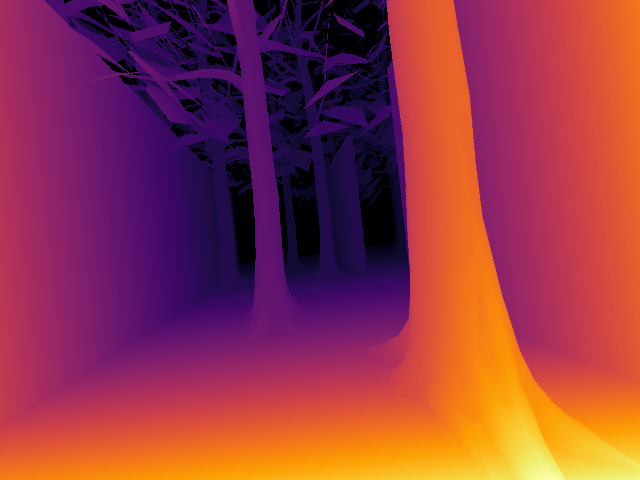}
    };
    \draw[anchor=south west] (10.47,0) node [text=white] {\small t~=~\SI{1.48}{\second}};

    \node[anchor=south west,inner sep=0] at (14.13,2.76) {
    \includegraphics[width=0.204\textwidth]{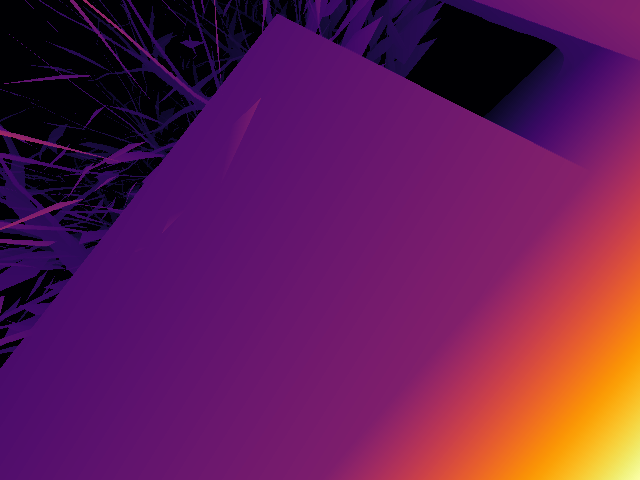}
    };
    \draw[anchor=south west] (14.13,2.76) node [text=white] {\small t~=~\SI{0.62}{\second}};
    \end{tikzpicture}   
      \captionof{figure}{
      Our approach focuses on learning navigation policies that map depth images to control commands. The task: flying to a goal in minimum time by navigating through obstacles. 
      (Left) We conducted an experiment using hardware-in-the-loop simulation, where the vehicle was flown in a real indoor arena while camera images were simultaneously rendered in real-time from a virtual-reality environment containing obstacles such as trees and walls.
      (Right) Four synthetic depth images rendered during flight at different time steps. 
    }
    \label{fig: teaser_img}
    \end{center}
    }
    ]
}

\definecolor{somegray}{rgb}{0.5, 0.5, 0.5}
\newcommand{\darkgrayed}[1]{\textcolor{somegray}{#1}}
\makeatletter
\newcommand*\titleheader[1]{\gdef\@titleheader{#1}}
\AtBeginDocument{%
  \let\st@red@title\@title
  \def\@title{%
    \vskip-4em
    \bgroup\normalfont\large\centering\@titleheader\par\egroup
    \vskip1.5em\st@red@title}
}

\titleheader{\darkgrayed{This paper has been accepted for publication at the
\\ IEEE International Conference on Robotics and Automation (ICRA), London 2023. 
\copyright IEEE} }

\title{\LARGE \bf
Learning Perception-Aware Agile Flight \\ in Cluttered Environments
}

\author{
    Yunlong Song$^{\ast}$$^{1}$, Kexin Shi$^{\ast}$$^{1}$, Robert Penicka$^{2}$, and Davide Scaramuzza$^{1}$ 
    \thanks{
     $^{\ast}$These two authors contributed equally. 
     $^{1}$Y. Song, K. Shi, and D. Scaramuzza are with the Robotics and Perception Group, Dep. of Informatics, University of Zurich, and Dep. of Neuroinformatics, University of Zurich and ETH Zurich, Switzerland (\protect\url{http://rpg.ifi.uzh.ch}).
    $^{2}$R. Penicka is with the Multi-robot Systems Group, Czech Technical University in Prague, Czech Republic.
    This work was supported by the Swiss National Science Foundation (SNSF) through the National Centre of Competence in Research (NCCR) Robotics, the Czech Science Foundation (GAČR) under research projects No. 23-06162M, the European Union’s Horizon 2020 Research and Innovation Programme under grant agreement No. 871479 (AERIAL-CORE), and the European Research Council (ERC) under grant agreement No. 864042 (AGILEFLIGHT).
    }
}

\begin{document}

\maketitle
\thispagestyle{empty}
\pagestyle{empty}

\input{Sections/abstract}

\input{Sections/introduction}
\input{Sections/relatedwork}
\input{Sections/methodology}
\input{Sections/experiments}
\input{Sections/conclusion}

\newpage

\balance

\bibliographystyle{IEEEtran}
\typeout{} 
\bibliography{references}

%
\end{document}

%% file: Sections/abstract.tex
\begin{abstract}
Recently, neural control policies have outperformed existing model-based planning-and-control methods for autonomously navigating quadrotors through cluttered environments in minimum time. 
However, they are not perception aware, a crucial requirement in vision-based navigation due to the camera's limited field of view and the underactuated nature of a quadrotor. 
We propose a learning-based system that achieves perception-aware, agile flight in cluttered environments. Our method  combines imitation learning with reinforcement learning (RL) by leveraging a privileged learning-by-cheating framework.
Using RL, we first train a perception-aware teacher policy with full-state information to fly in minimum time through cluttered environments.
Then, we use imitation learning to distill its knowledge into a vision-based student policy that only perceives the environment via a camera.
Our approach tightly couples perception and control, showing a significant advantage in computation speed ($10\times$ faster) and success rate. 
We demonstrate the closed-loop control performance using hardware-in-the-loop simulation.

\vspace{4mm}

Video: \url{https://youtu.be/9q059CFGcVA} 
\end{abstract}

%% file: Sections/introduction.tex
\section{INTRODUCTION}
Vision-based navigation of micro aerial vehicles has recently achieved impressive results outside of research labs, from exploring Mars to swarm navigation~\cite{zhou2022swarm} and agile flight in the wild~\cite{loquercio2021wild}. 
%
%
However, existing methods focused on the general task of reaching a goal while navigating in cluttered unknown environments and much less on reaching the goal in \emph{minimum time} (also known as time-optimal flight).
Additionally, in many scenarios, such as a search-and-rescue and reconnaissance in a known  environment, these approaches are not ideally suited as they do not leverage \emph{prior information} about the environment that might be available. 

In this work, we tackle the problem of vision-based, \emph{minimum-time} flight in cluttered, \emph{known} environments for quadrotor drones.
Minimum-time flight requires the vehicle to operate on the edge of its physical limits (high speeds and accelerations) and perceptual limits (limited field of view, motion blur, limited sensing range, fast reaction times). 
The limited field of view of the onboard camera is particularly constraining for quadrotors due to their underactuated nature: in the most common configuration, all the rotors point in the same direction, which causes the robot to accelerate only in this direction. If the camera is rigidly attached to the drone, this means that a trade-off must be found between maximizing flight performance and optimizing the visibility of regions of interest. We refer to this problem as \emph{perception-aware flight}~\cite{Falanga17ICRA,falanga2018pampc}.


Recently, reinforcement learning-based methods~\cite{song2021autonomous, penicka2022learning} have been proposed in order to address the planning and control problem for minimum-time flight.
In particular, \cite{penicka2022learning} presented a neural network controller trained via reinforcement learning that outperformed previous model-based planning-and-control methods in cluttered environments. 
Their controller maps ground-truth state observations directly to control commands, forgoing the need for a high-level trajectory planner and significantly reducing potential compounding errors and overall system latency. 
%
%
However, the learned controller is optimized only for maximizing the progress along a reference path while avoiding obstacles, and completely ignores the perception constraint induced by the camera's field of view.
As a result, there is no guarantee of visibility of regions of interest (i.e., no perception awareness).


\rebuttal{
We propose a vision-based navigation system to fly a quadrotor through cluttered environments at high-speed with perception awareness. Our method  combines imitation learning and reinforcement learning (RL) by leveraging a privileged learning-by-cheating framework.
We begin by training a state-based teacher policy using deep RL to fly a minimum-time trajectory in cluttered environments. This policy integrates progress maximization and obstacle avoidance with a perception-aware reward that aligns the camera orientation with the flight direction. Next, by imitating the teacher policy, we train a vision-based policy that does not rely on privileged information about the obstacles. The resulting vision-based policy achieves high-speed flight and high success rates. We show that our policy has very low computational latency (just~\SI{1.4}{\milli\second}) compared to classical methods with intermediate map representations that have 10 times higher latency.

To validate our approach, we test the closed-loop control performance of our vision-based policy in the real world using hardware-in-the-loop (HITL) simulation. HITL involves flying a physical quadrotor in a motion-capture system while observing virtual photorealistic environments that are updated in real-time. Unlike purely synthetic experiments, HITL simulation employs real-world dynamics and proprioceptive sensors while allowing us to render arbitrarily dense and complex environments without risking a drone crash. Our vision-based policy transfers to the real world despite system delays and a mismatch of the vehicle model.}

%% file: Sections/relatedwork.tex
\section{RELATED WORK}

Various approaches have been studied in the literature for vision-based agile flight in cluttered environments. 
Particularly, in a traditional robotics design, the navigation task is divided into mapping, planning, and control. 
This line of work first requires computationally-intensive algorithms, such as Simultaneous Localization and Mapping~(SLAM), to infer the 3D structure of the environment from 2D noisy image data~\cite{zhou2022swarm, heng2014autonomous, scaramuzza2014vision, blosch2010vision}. 
Given a 3D map of an environment, planning algorithms generate 
feasible trajectories that follow the shortest collision-free path from start to goal utilizing the vehicle's full dynamic capabilities. 
Some planning algorithms exploit the effective differential flatness of
the platform using polynomial or B-spline representations~\cite{mellinger2011minimum, Richter2013ISRR}, whereas others rely on nonlinear programming~\cite{foehn2021time} or searching-based planning~\cite{liu2018search}. 
Finally, a controller is used to follow the trajectory precisely, such as model predictive control~\cite{falanga2018pampc, song2022policy} or differential-flatness-based control~\cite{faessler2017differential}. 


Dividing the navigation task into a sequence of subtasks allows for simplifying the problem and parallelizing each component's development, resulting in an interpretable system from the engineering perspective. 
However, it leads to a pipeline that is sensitive to unmodeled effects
due to a lack of interactions between each component. 
Also, the system requires additional latency for passing or waiting for the information.

Recently, learning-based methods attempted to address the aforementioned limitations. 
For instance, researchers~\cite{loquercio2021wild} propose to directly map noisy sensory observations to collision-free trajectories in a receding-horizon fashion. 
This direct mapping forgoes the need for 3D environment mapping and collision waypoints planning. 
Though it reduces processing latency and increases robustness to noisy perception, a controller is still required to track the trajectory. 
Another approach\cite{adamkiewicz22ral} leverages recent advances in neural radiance fields (NeRF)~\cite{mildenhall2021nerf} to navigate a drone through a pre-mapped 3D environment using only a monocular camera.
However, the navigation approach requires the offline construction of a NeRF for each new target environment, which can be computationally expensive. 

Some other works propose to learn end-to-end policies directly from sensory observations to control commands using imitation learning~\cite{ross2013learning, sadeghi2016cad2rl, zhang2016learning}. 
Without relying on an expert, deep reinforcement learning has the potential to find more optimal policies for a variety of tasks.
For aerial robots, Deep RL was successfully applied to multiple end-to-end visual navigation tasks such as object following~\cite{sampedro18icra}, exploration~\cite{devo2022ral}, and obstacle avoidance~\cite{singla21its, kim22esa}.

%% file: Sections/methodology.tex
\section{METHODOLOGY}
\begin{figure*}[!htbp]
\centering
\includegraphics[scale=0.28]{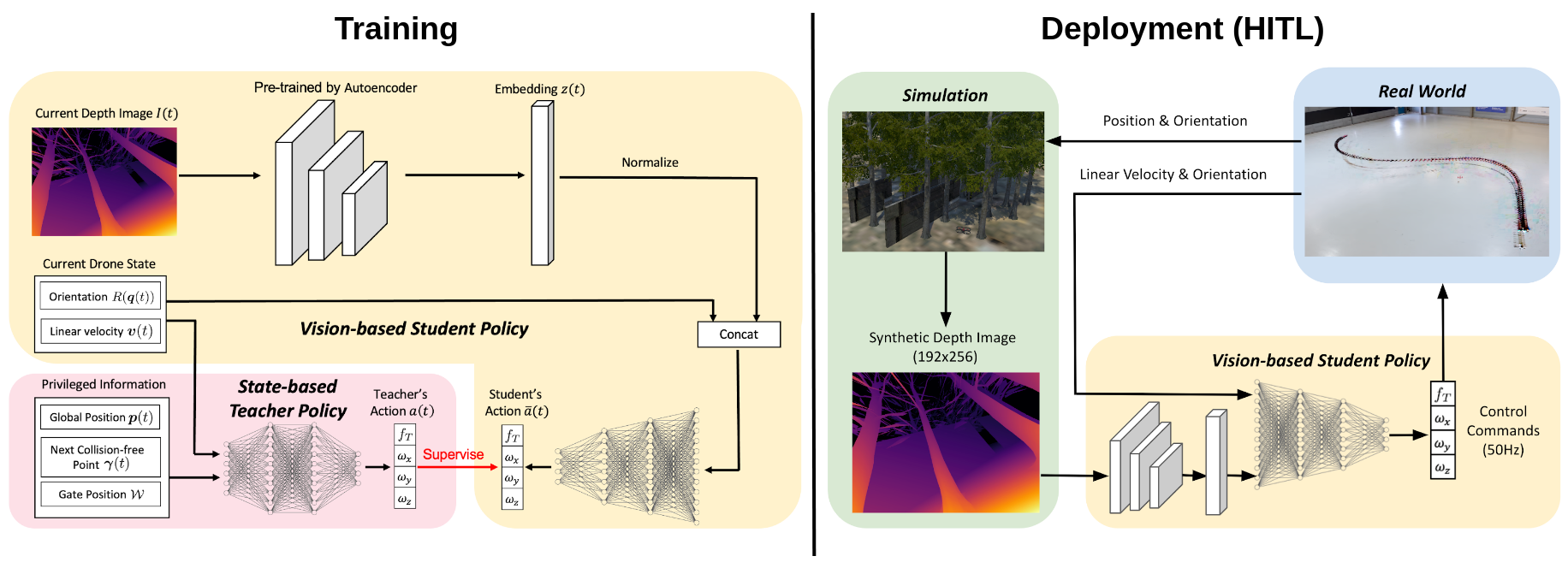}
\caption{\textbf{System overview.} First, we train a state-based teacher policy with access to privileged state information. Then we distill the teacher's knowledge into a vision-based student policy. 
After finishing training in simulation, we directly deploy the vision-based student policy in the real world via hardware-in-the-loop~(HITL) simulation.}
\label{overview}
\end{figure*}

The main goal of the presented controller is to fly a quadrotor through cluttered environments as fast as possible. 
The quadrotor observes the environment using a depth camera.
Our policy controls the vehicle directly from the sensory observation. 

\subsection{Method Overview}
Figure~\ref{overview} shows an overview of the system. 
We use a simple privileged reinforcement learning framework to tackle high-dimensional observations. 
We first train a state-based teacher policy using model-free deep reinforcement learning, in which the policy has access to privileged information about the vehicle's state and its surrounding environment.
This teacher policy is then distilled into a vision-based student policy that does not rely on privileged information. 


\subsection{Learning State-based Teacher Policy}
Our teacher policy is a two-layer multilayer perceptron (MLP) that can achieve minimum-time flight in a cluttered environment~\cite{penicka2022learning}. 
The key idea of learning a teacher policy for minimum-time flight in cluttered environments is three-fold: 1) generation of a
topological guiding path using a probabilistic roadmap method,
2) optimization of a task objective that combines progress maximization along the guiding path with obstacle avoidance, and
3) a curriculum training strategy to train a neural network
policy using deep reinforcement learning.
At each time step, the policy directly controls the vehicle given 
an observation about the quadrotor's full-state, a collision-free reference waypoint, and a collision-free reference line segment. 
Learning a state-based policy is fast since it does not require rendering images for training.


\textbf{Perception-aware Obstacle Avoidance:}
However, the previously proposed objective neglects the camera orientation, which is crucial for vision-based flight. 
Due to the camera's limited field of view, it is imperative to align the camera orientation with the flight direction. 
For instance, when navigating through an unknown environment, maximizing the visibility of the next waypoint allows for counteracting unexpected scenarios. 
To this end, we combine the minimum-time obstacle avoidance objective with a perception-aware reward. 
The perception-aware reward is formulated as 
\begin{equation}
 r_{pa} = \exp(-\Vert{\boldsymbol{\theta_{yaw}} - \boldsymbol{\theta_{dir}}}\Vert),
\end{equation}
where $\boldsymbol{\theta_{yaw}}$ is the current yaw angle of camera and $\boldsymbol{\theta_{dir}}$ is the next flight direction defined by $\boldsymbol{\gamma}(t)-\boldsymbol{p}(t)$. 
\rebuttal{
Here, $\boldsymbol{\gamma}(t)$ is the farthest collision-free point on the reference path, namely, the vector $\boldsymbol{\gamma}(t)-\boldsymbol{p}(t)$ does not intersect with the obstacles while maintaining longest segment length. 
It indicates both the flight direction and path length for the quadrotor to maximize progress. }
Finally, the stage reward is denoted as
\begin{align*}
r(t) = & k_p r_p(t) + k_s s(\boldsymbol{p}(t)) + k_{wp} r_{wp} \\
    & + k_{obs} r_{obs} + k_\omega\|\boldsymbol{\omega}\| + k_{pa} r_{pa},
\end{align*}
where $r_p(t)$ is the progress reward, $s(\boldsymbol{p}(t))$ is the reached distance reward, $r_{wp}$ is the reward for passing through a corresponding waypoint, $r_{obs}$ is the penalty for collision with obstacles, $\|\boldsymbol{\omega}\|$ is the penalty for large angular velocity, $r_{pa}$ is the perception-aware reward. 
In sparse environments (including Columns, Office, Racing, and Real Flight)~(Sec.~\ref{sec:exp}), the reward coefficients $k_p$, $k_s$, $k_{wp}$, $k_{obs}$, $k_\omega$ and $k_{pa}$ are 5.0, 0.01, 5.5, -0.5, -0.02 and 0.05 respectively. In the challenging Racing MW environment, $k_{pa}$ is set as 0.1 while keeping other coefficients the same.


\subsection{Learning Vision-based Student Policy}
We use the trained teacher policy to supervise a vision-based student policy. 
Our student policy removes the assumption about perfect state information of the obstacles and a reference collision-free line segment. 
Instead, the student policy reacts to obstacles purely based on the current observation from a depth camera. 
Specifically, our student policy extracts low-dimensional feature embeddings of the depth image using a CNN autoencoder.
Given the embeddings, together with the current vehicle's orientation and velocity, an MLP is used to regress the control commands. 
%

%

To effectively reduce the dimension of depth images, we train an autoencoder\cite{hinton2006reducing} to learn low-dimensional feature representations for the depth image. 
%
%
The encoder $\boldsymbol{E_\phi}$ contains three convolution layers with decreasing image size and increasing receptive field gradually. 
Then a decoder $\boldsymbol{D_\theta}$, which is a three-layers MLP, is used to reconstruct the original images. 
We train this autoencoder using depth images collected by the teacher policy. 
A standard L2 loss is used for training the autoencoder:
\begin{equation}
L(\theta, \phi)=\frac{1}{T} \sum_{t=1}^T\left\|\boldsymbol{I}(t)-\boldsymbol{D_\theta}\left(\boldsymbol{E_\phi}\left(\boldsymbol{I}(t)\right)\right)\right\|_2^2.
\end{equation}
After we obtain the optimal paremeter $\phi$ of encoder, we can represent the depth images by its latent embedding $\boldsymbol{z}(t) = \boldsymbol{E_\phi}(\boldsymbol{I}(t))$. 

We use the pre-trained autoencoder to extract low-dimensional feature representations for the input depth image. 
The image embedding $\boldsymbol{z}(t)$ generated by the encoder is normalized since it is beneficial for convergence empirically.
We construct an observation vector by concatenating the image embedding with the vehicle's partial states, including linear velocity $\boldsymbol{v}(t)$ and orientation $R(\boldsymbol{q}(t))$.
The observation vector is then fed into an MLP to regress student's action. 
We define the action loss as 
\begin{equation}
L(\psi)=\frac{1}{T} \sum_{t=1}^T\left\|\boldsymbol{a}(t) - \boldsymbol{\overline{a}}(t)| \psi\right\|_2^2
\end{equation} 
to minimize the difference between teacher's action $\boldsymbol{a}(t)$ and the output of student policy $\boldsymbol{\overline{a}}(t)$, where $\psi$ is the parameter of student policy. 
Once we gain the optimal parameter $\psi$, the vision-based student can imitate the state-based teacher's behavior to the maximum extent. 

\subsection{Hardware-in-the-loop Simulation}
Developing vision-based navigation algorithms for minimum-time agile flight is not only time-consuming but also expensive. 
This is due to the large amount of data required for training and testing perception algorithms in diverse environments, some of which can be even harmful or risky to humans, such as collapsed buildings. 
It progressively becomes less safe and more expensive since more aggressive flights can lead to devastating crashes.

We propose hardware-in-the-loop~(HITL) simulation for evaluating vision-based policies using real-world physics and virtual photorealistic environments. 
HITL simulation combines the advantage of both testing with a physical platform and rendering diverse testing scenarios in an
inexpensive manner. 

We use a motion capture system to capture accurate state information about the vehicle and simultaneously simulate the vehicle's motion in any virtual unstructured environment with arbitrary complexity. 
The policy has to control the physical drone using synthetic images while handling sim-to-real gaps introduced by the physical system, such as delay and vibration. 
When the policy experiences failures, such as flying toward the ground, a safety guard that is based on a state-based controller is triggered.
Hence, the drone is not damaged by virtual obstacles and is safe when the policy makes false decisions. 

%% file: Sections/experiments.tex
\section{Experiments}
\label{sec:exp}

We evaluate our method in both controlled simulation environments and the real world. 
We first compare the proposed vision-based policy against methods based on planning-and-control and the state-based policy.
Second, we study the computational latency for our policy and show a comparison against 
state-of-the-art methods in collision avoidance.
We verify our vision-based policy in the real world using a high-performance racing drone and hardware-in-the-loop simulation.

\textbf{Policy Training:}
We train our neural network policy using the Flightmare~\cite{yunlong2020flightmare} simulator.
We use a customized implementation of the proximal policy optimization algorithm (PPO)~\cite{schulman2017proximal} to train the teacher policy.
We implement a CNN autoencoder to learn feature representations from depth images and an imitation learning pipeline to distill the teacher's knowledge into a vision-based student policy,  

\textbf{Simulation:}
We perform a set of controlled simulation experiments to compare our methods' performance with several state-of-the-art baselines. 
For benchmark comparison, we use the same three environments from~\cite{penicka2022learning}, denoted as Columns, Office, and Racing.
We use two performance criteria for the evaluation: (1) average flight time from a given starting point to the goal position and (2) success rate.
The success rate is defined by starting the policy from different starting positions drawn from a uniform distribution with 20 runs.
The results of these experiments are summarized in Table~\ref{tab:state_vision}.

\begin{figure*}[!htb]
   \centering
   \setlength\tabcolsep{-0.5pt}
  \begin{tabular}{cccc}
  \subcaptionbox{Columns\label{fig:columns}}{
    \includegraphics[width=0.32\linewidth]{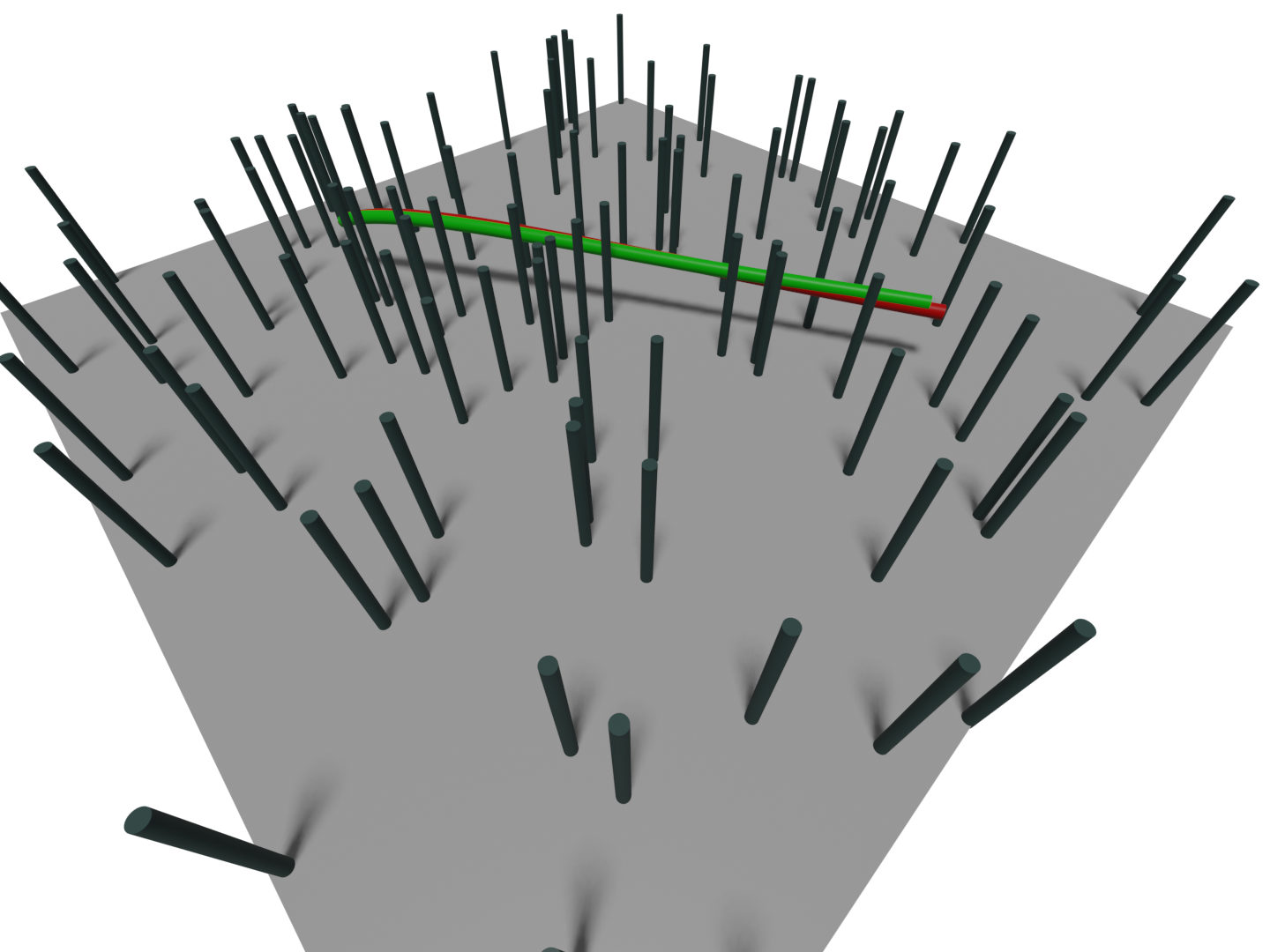}
    \vspace{-0.5em}
  }   &  
  \subcaptionbox{Office\label{fig:office}}{
    \includegraphics[width=0.32\linewidth]{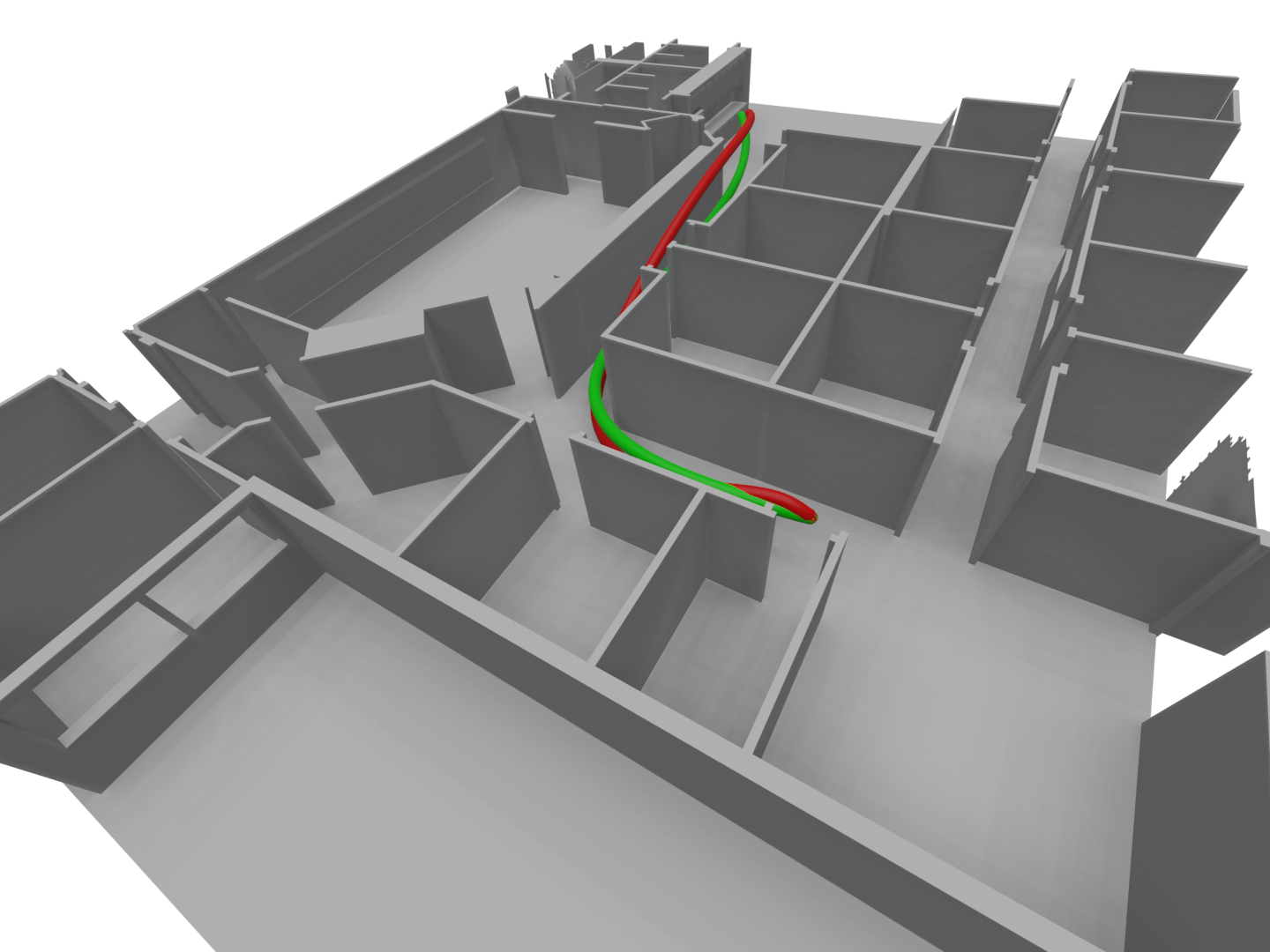}
    \vspace{-0.5em}
  }   
    &
  \subcaptionbox{Racing / Racing MW\label{fig:racing}}{
 \includegraphics[width=0.32\linewidth]{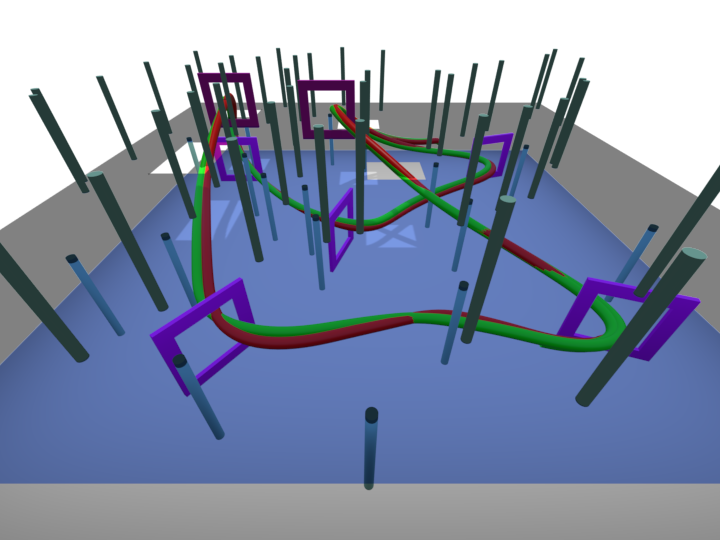}
 \vspace{-0.5em}
  }   
 \end{tabular}
   \vspace{-0.1em}
   \caption{
   \textbf{An overview of three environments used for baseline comparisons.}
   The curves show two trajectories flown by the state-based teacher policy (red) and the vision-based student policy (green). 
    \label{fig:environments} 
    }
\end{figure*}

%

Furthermore, in terms of computation latency, we compare our policy against a mapping-based method~\cite{zhou2019robust}, a reactive method~\cite{florence2020integrated}, and a learning-based method~\cite{loquercio2021wild}. 
We define computation latency as the time required to pre-process the capture depth images and to generate the control commands. 
The computation latency plays an important role when flying at high speed due to the limited sensing range of a physical camera~\cite{aFalanga19ral_howfast}. 

\textbf{Real world:}
We deploy the vision-based policy in the real world using a high-performance racing drone and
the Agilicious~\cite{foehn2022agilicious} control stack. 
We design a lightweight quadrotor platform that has a total weight of 540 grams and can produce a maximum thrust of about 34~N, which
results in a thrust-to-weight ratio around 6:1.
We conduct our experiments in one of the world’s largest indoor drone-testing arenas ($30 \times 30 \times 8$~m$^3$) equipped with a motion capture system with an operating frequency of up to~\SI{400}{\hertz}.

\subsection{Baseline Comparisons}
\label{sec: baseline_comparisons}
This section shows that a vision-based policy can navigate the vehicle in cluttered environments reliably without privileged information about the state of the vehicle and the obstacles. 
We compare the proposed vision-based perception-aware student policy with the non-perception-aware sampling-based method~\cite{penicka2022minimum} and a state-based data-driven method~\cite{penicka2022learning}. 
We use three different environments (Columns, Office, Racing), and in each, we test four cases with different starting points and goal positions. 
In the Racing environment, we additionally set multiple waypoints (Racing MW) for the drone to pass through in order to increase the difficulty of the track. 

\input{tables/obs_raw_perform.tex}

Table~\ref{tab:state_vision} shows the comparison results. 
The sampling-based method~\cite{penicka2022minimum} achieves overall very low success rates.
This is because the planned time-optimal trajectory by definition pushes the vehicle to its maximum performance. 
When tracking such a trajectory using an obstacle-blind controller, such as model predictive control~(MPC), the controller struggles to follow the trajectory due to diminishing control authority. 
As a result, such a planning-and-control pipeline is very sensitive to unknown disturbances and model mismatches. 
%
%
On the contrary, data-driven methods can achieve a high success rate by simulating diverse scenarios during training, including random disturbances, random initial starting points, and random physical parameters. 

\rebuttal{
Our vision-based policy achieves the same success rates as the state-based policy without relying on privileged information about the obstacle. However, due to minor action errors, the student's policy tends to exhibit \emph{undesired} riskier behaviors compared to the teacher's policy. Specifically, the trajectory flown by the student policy often cuts corners, resulting in a shorter path and faster lap time, but increased risks.
To minimize the risk, we employ a conservative distance margin; it implies that the distance maintained between the vehicle and the obstacle is greater than the minimum required distance to avoid collisions. Hence, the student policy can still achieve high success rates. 
Finally, we can see that the perception-aware policy does not have to sacrifice flight time compared to the non-perception-aware state-based policy~\cite{penicka2022learning}, with the exception of the Racing MW environment.
In this specific scenario, the highly cluttered environment necessitates compromising the progress maximization objective with the alignment of the camera's orientation, resulting in a slower, but perception-aware, policy.
}






\subsection{Computational Latency}
Our policy manifests a significant advantage in achieving low computational latency.
Table~\ref{tab:latency} shows a comparison of the computational latency between our method and baseline methods. The latency is obtained from~\cite{loquercio2021wild}.  
The FastPlanner~\cite{zhou2019robust} method experiences the highest computational latency, mainly due to mapping. 
The reactive method~\cite{florence2020integrated} can significantly reduce the computation latency by taking out the mapping component. 
The data-driven method~\cite{loquercio2021wild} achieves a much faster computation speed with a total latency of only~\SI{10.3}{\milli\second}. 
Depending on which controller is used for tracking the planned trajectory, additional latency caused by the controller should also be considered in the baseline methods.
For instance, trajectory tracking using a differential-flatness-based control has a computation time in the magnitude of only microseconds, while nonlinear MPC requires several milliseconds~\cite{Sun2021tro}. 
Finally, our reactive vision-based policy achieves the fastest computation time of only around~\SI{1.41}{\milli\second} for conducting perception, planning and control jointly. 

\input{tables/latency.tex}



\subsection{Hardware-in-the-Loop Flight}
\begin{figure}[htp]
\centering
\includegraphics[width=0.4\textwidth]{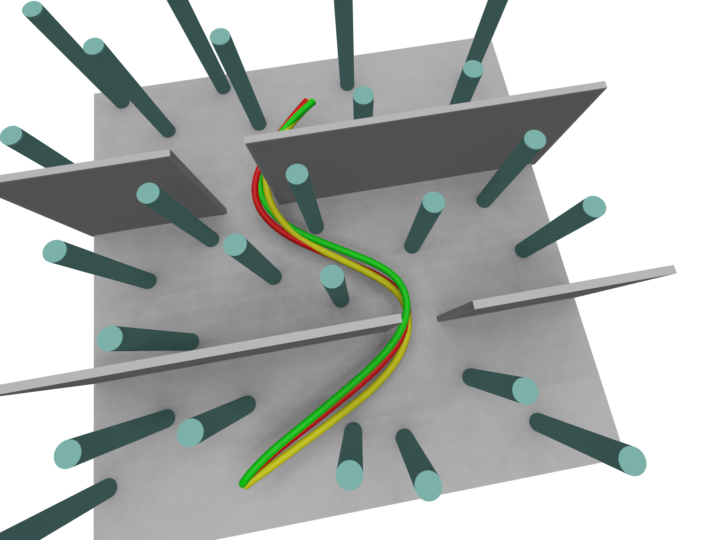}
\caption{Vehicle's trajectories from the real world (yellow), from simulation with teacher policy (red), and simulation with student policy (green).}
\label{fig:realworld_traj}
\end{figure}

To test the closed-loop control performance of our neural network policy in the real world,
we deploy our vision-based policy using hardware-in-the-loop simulation. 
A time-lapse illustration of the real-world flight is shown in Figure~\ref{fig: teaser_img}, where we merge the simulated forest environment with the real-world arena for visualization purposes.
Fig.~\ref{fig:realworld_traj} shows a visualization of the real-world trajectory (yellow) and the simulation trajectories by the teacher policy (green) and the student policy (red).
All trajectories are collision-free and have similar flight time. 
Our policy achieves a maximum speed of~\SI{54.36}{\kilo\meter\per\hour} in the real world using offboard control, in which the control commands are computed using a workstation and sent to the drone via a remote transmitter. 
We achieved a control frequency of~\SI{50.0}{\hertz}; this is possible because our system has, on average, a total computational latency of only around~\SI{1.41}{\milli\second}.

\subsection{On the Importance of Perception-Aware Flight}

\rebuttal{
We conduct an ablation study to evaluate the impact of perception awareness on vision-based flight in the challenging Racing MW environment. The results are presented in Table III. We observed that the vehicle achieved faster lap times (7.78~\SI{}{\second}) when flying non-perception-aware since the policy only prioritized maximizing progress along the path without considering camera orientation. In contrast, perception-aware flight led to higher lap times (8.52~\SI{}{\second}) since the policy had to balance maximizing progress and aligning drone orientation to the next flight direction.

However, the non-perception-aware vision-based policy achieved a lower success rate because the camera orientation was not well-aligned with the flight direction, making it difficult to respond to upcoming obstacles in time. As discussed in Section~\ref{sec: baseline_comparisons}, due to action errors, the student policy trajectory often cut corners, resulting in a shorter path (faster lap time) than the teacher policy but increased risk.}

\input{tables/ablation}

%% file: tables/obs_raw_perform.tex
\begin{table*}[htp]
\renewcommand\arraystretch{1.2}
\renewcommand{\tabcolsep}{6.4pt} 
\centering
\caption{Comparison between the baseline algorithm~\cite{penicka2022minimum}  and the proposed state-based and vision-based policies in non- and perception-aware mode. 
\label{tab:state_vision}}
\begin{tabular}{lccccccccc}
\hline
\multirow{4}*{\textbf{Environment}} & \multirow{4}*{\textbf{Case}} & 
\multicolumn{4}{c}{\textbf{Non-perception-aware}} & \multicolumn{4}{c}{\textbf{Perception-aware (ours)}}\\
\cmidrule(lr){3-6} \cmidrule(lr){7-10}
& & \multicolumn{2}{c}{\textbf{Sampling-based~\cite{penicka2022minimum} + MPC~\cite{Foehn2021science}}} & \multicolumn{2}{c}{\textbf{State-based~\cite{penicka2022learning}}} & \multicolumn{2}{c}{\textbf{State-based (teacher)}} & \multicolumn{2}{c}{\textbf{Vision-based (student)}}  \\
\cmidrule(lr){3-4} \cmidrule(lr){5-6} \cmidrule(lr){7-8} \cmidrule(lr){9-10} 
                        &   & Success[\%] & Time[s] & Success[\%] & Time[s] & Success[\%] & Time[s]     & Success[\%]      & Time[s] \\
\hline
\multirow{4}*{Columns}   & 0 & 25 & 1.22     & 100        & \textbf{1.10$\pm$0.02} & 100        & 1.16$\pm$0.02        & 100      & \textbf{1.10$\pm$0.02} \\
                        & 1 & 0 & -         & 100        & \textbf{1.10$\pm$0.02} & 100        & \textbf{1.10$\pm$0.03}        & 100      & \textbf{1.10$\pm$0.03} \\
                        & 2 & 27 & 1.14     & 100        & \textbf{1.10$\pm$0.02} & 100        & \textbf{1.10$\pm$0.02}        & 100      & \textbf{1.10$\pm$0.03} \\
                        & 3 & 16 & 1.70     & 100        & 1.40$\pm$0.04 & 100        & 1.46$\pm$0.04        & 100      & \textbf{1.39$\pm$0.04}  \\
\hline
\multirow{4}*{Office}   & 0 & 41 & 2.38     & 100        & 1.80$\pm$0.04 & 100        & 1.92$\pm$0.04        & 100      & \textbf{1.79$\pm$0.04}  \\
                        & 1 & 28 & 1.86     & 100        & 1.80$\pm$0.03 & 100        & 1.82$\pm$0.02        & 100      & \textbf{1.76$\pm$0.02}  \\
                        & 2 & 56 & 2.29     & 100        & \textbf{1.62$\pm$0.02} & 100        & 1.66$\pm$0.03        & 100      & 1.63$\pm$0.03  \\
                        & 3 & 70 & 2.16     & 100        & 1.46$\pm$0.02 & 100        & 1.48$\pm$0.02        & 100      & \textbf{1.45$\pm$0.02}  \\
\hline
\multirow{4}*{Racing}   & 0 & 57 & 1.61     & 100        & 1.33$\pm$0.03 & 100        & \textbf{1.32$\pm$0.02}        & 100      & 1.33$\pm$0.03  \\
                        & 1 & 51 & 1.64     & 100        & \textbf{1.30$\pm$0.02} & 100        & 1.42$\pm$0.03        & 100      & 1.31$\pm$0.04  \\
                        & 2 & 76 & 1.72     & 100        & \textbf{1.42$\pm$0.02} & 100        & 1.44$\pm$0.02        & 100      & 1.43$\pm$0.02  \\
                        & 3 & 54 & 1.80     & 100        & 1.42$\pm$0.02 & 100        & 1.48$\pm$0.02        & 100      & \textbf{1.39$\pm$0.01}  \\
\hline
Racing MW             & 0 & 25 & 7.22\footnote{This result was obtained using the same track, but without looping.}       & 100        & \textbf{7.78$\pm$0.07}  & 100        & 8.52$\pm$0.04        & 100      &  8.34$\pm$0.06\\
\hline
\end{tabular}
\end{table*}

%% file: tables/latency.tex
\begin{table}[htp]
\centering
{
\renewcommand\arraystretch{1.2}
\renewcommand{\tabcolsep}{4.0pt} 
\caption{Comparison of planning and control latency.}
\begin{tabular}{l|ccccc}
\toprule
\multirow{2}{*}{\textbf{Method}} & \multirow{2}{*}{\textbf{Components}} & \multirow{2}{*}{\textbf{$\mu$[ms]}} & \multirow{2}{*}{\textbf{$\sigma$[ms]}} & \multirow{2}{*}{\textbf{Prec.[\%]}} & \textbf{Total Proc.} \\
& & & & & \textbf{Latency[ms]} \\
\hline
\multirow{3}{*}{\thead[l]{FastPlanner\\ \cite{zhou2019robust}}}   & Pre-processing & 14.6  & 2.3 & 22.3 & \multirow{3}*{65.2} \\
                             & Mapping & 49.2 & 8.7 & 75.5 & \\
                             & Planning & 1.4 & 1.6 & 2.2 & \\
\hline
\multirow{2}{*}{\thead[l]{Reactive\\ \cite{florence2020integrated}}}      & Pre-processing & 13.8 & 1.3 & 72.3 &  \multirow{2}*{19.1}\\
                             & Planning & 5.3  & 0.9 & 27.7 &   \\
\hline
 \multirow{3}{*}{\thead[l]{Agile\\autonomy\\ \cite{loquercio2021wild}}} & Pre-processing & 0.1  & 0.04 & 3.9 & \multirow{3}*{10.3} \\
                             & NN inference & 10.1 & 1.5 & 93.0 & \\
                             & Projection & 0.08 & 0.01 & 3.1 & \\
\hline
\multirow{2}{*}{ \textbf{Ours}}      & Pre-processing & 0.23 & 0.04 & 16.3 & \multirow{2}*{\textbf{1.41}} \\
                             & NN inference & 1.18 & 0.08 & 83.7 &   \\
\bottomrule
\end{tabular}
\label{tab:latency}
}
\end{table}

%% file: tables/ablation.tex
\begin{table}[htp]
\centering
{
\renewcommand\arraystretch{1.2}
\renewcommand{\tabcolsep}{4.0pt} 
\caption{Ablation study of the perception-aware reward.}
\begin{tabular}{cc|cc}
\toprule
 &  & Success[\%] & Time[\SI{}{\second}]\\
\hline
\multirow{2}{*}{\thead[l]{\textbf{Non-perception-aware}\\}}   & \textbf{State-based} & 100 & 7.78$\pm$0.07 \\
                                                            & \textbf{Vision-based} & 70 & 7.66$\pm$0.05 \\
\hline
\multirow{2}{*}{\thead[l]{\textbf{Perception-aware}\\}}   & \textbf{State-based} & 100 & 8.52$\pm$0.04 \\
                                                            & \textbf{Vision-based} & 100 & 8.34$\pm$0.06 \\
\bottomrule
\end{tabular}
\label{tab:ablation}
}
\end{table}

%% file: Sections/conclusion.tex
\section{Discussion}
We have observed limitations in our policy's robustness to random disturbances and its generalizability to other environments beyond its original training. We have identified two main design choices as the cause: the neural network architecture and the training pipeline.

First, our control policy is solely reactive, lacking the capability to retain a memory or historical observations within the network. This shortcoming can be resolved through the use of more advanced policy representations, such as Long Short-Term Memory (LSTM)~\cite{hochreiter1997long} or Transformers~\cite{vaswani2017attention}. These architectures enable the network to effectively incorporate past experiences, leading to a more robust and adaptable control policy.
Second, our policy is trained using a fixed environment that does not account for sensor noises. To address this limitation and enhance the policy's robustness, domain randomization along with data augmentation can be leveraged in order to add variability and diversity to the training data, resulting in a more versatile policy that can better handle real-world scenarios with varying levels of sensor noise.

\section{Conclusion and Discussion}
\label{sec:con}
This paper presented a method to learn perception-aware neural network policies for vision-based agile flight in cluttered environments.
Our method leverages a privileged learning-by-cheating framework to achieve efficient training. 
\rebuttal{
As a result, our work demonstrates an end-to-end policy that can achieve strong performance in high-speed flight in cluttered environments while also providing significant computational advantages. 
Our results in the real world via HITL suggest that end-to-end policies represent a promising approach for enabling agile flight in challenging real-world scenarios.
}

%% file: main.bbl
\begin{thebibliography}{10}
\providecommand{\url}[1]{#1}
\csname url@samestyle\endcsname
\providecommand{\newblock}{\relax}
\providecommand{\bibinfo}[2]{#2}
\providecommand{\BIBentrySTDinterwordspacing}{\spaceskip=0pt\relax}
\providecommand{\BIBentryALTinterwordstretchfactor}{4}
\providecommand{\BIBentryALTinterwordspacing}{\spaceskip=\fontdimen2\font plus
\BIBentryALTinterwordstretchfactor\fontdimen3\font minus
  \fontdimen4\font\relax}
\providecommand{\BIBforeignlanguage}[2]{{%
\expandafter\ifx\csname l@#1\endcsname\relax
\typeout{** WARNING: IEEEtran.bst: No hyphenation pattern has been}%
\typeout{** loaded for the language `#1'. Using the pattern for}%
\typeout{** the default language instead.}%
\else
\language=\csname l@#1\endcsname
\fi
#2}}
\providecommand{\BIBdecl}{\relax}
\BIBdecl

\bibitem{zhou2022swarm}
X.~Zhou, X.~Wen, Z.~Wang, Y.~Gao, H.~Li, Q.~Wang, T.~Yang, H.~Lu, Y.~Cao, C.~Xu
  \emph{et~al.}, ``Swarm of micro flying robots in the wild,'' \emph{Science
  Robotics}, vol.~7, no.~66, p. eabm5954.

\bibitem{loquercio2021wild}
A.~Loquercio, E.~Kaufmann, R.~Ranftl, M.~Müller, V.~Koltun, and D.~Scaramuzza,
  ``Learning high-speed flight in the wild,'' \emph{Science Robotics}, vol.~6,
  no.~59, p. eabg5810, 2021.

\bibitem{Falanga17ICRA}
\BIBentryALTinterwordspacing
D.~Falanga, E.~Mueggler, M.~Faessler, and D.~Scaramuzza, ``Aggressive quadrotor
  flight through narrow gaps with onboard sensing and computing using active
  vision,'' in \emph{{IEEE} Int. Conf. Robot. Autom. (ICRA)}.\hskip 1em plus
  0.5em minus 0.4em\relax {IEEE}, 2017, pp. 5774--5781. [Online]. Available:
  \url{https://doi.org/10.1109/ICRA.2017.7989679}
\BIBentrySTDinterwordspacing

\bibitem{falanga2018pampc}
D.~Falanga, P.~Foehn, P.~Lu, and D.~Scaramuzza, ``Pampc: Perception-aware model
  predictive control for quadrotors,'' in \emph{IEEE/RSJ Int. Conf. Intell.
  Robot. Syst. (IROS)}.\hskip 1em plus 0.5em minus 0.4em\relax IEEE, 2018, pp.
  1--8.

\bibitem{song2021autonomous}
Y.~Song, M.~Steinweg, E.~Kaufmann, and D.~Scaramuzza, ``Autonomous drone racing
  with deep reinforcement learning,'' in \emph{IEEE/RSJ Int. Conf. Intell.
  Robot. Syst. (IROS)}, 2021.

\bibitem{penicka2022learning}
R.~Penicka, Y.~Song, E.~Kaufmann, and D.~Scaramuzza, ``Learning minimum-time
  flight in cluttered environments,'' \emph{IEEE Robotics and Automation
  Letters}, vol.~7, no.~3, pp. 7209--7216, 2022.

\bibitem{heng2014autonomous}
L.~Heng, D.~Honegger, G.~H. Lee, L.~Meier, P.~Tanskanen, F.~Fraundorfer, and
  M.~Pollefeys, ``Autonomous visual mapping and exploration with a micro aerial
  vehicle,'' \emph{Journal of Field Robotics}, vol.~31, no.~4, pp. 654--675,
  2014.

\bibitem{scaramuzza2014vision}
D.~Scaramuzza, M.~C. Achtelik, L.~Doitsidis, F.~Friedrich, E.~Kosmatopoulos,
  A.~Martinelli, M.~W. Achtelik, M.~Chli, S.~Chatzichristofis, L.~Kneip
  \emph{et~al.}, ``Vision-controlled micro flying robots: from system design to
  autonomous navigation and mapping in gps-denied environments,'' \emph{IEEE
  Robotics \& Automation Magazine}, vol.~21, no.~3, pp. 26--40, 2014.

\bibitem{blosch2010vision}
M.~Bl{\"o}sch, S.~Weiss, D.~Scaramuzza, and R.~Siegwart, ``Vision based mav
  navigation in unknown and unstructured environments,'' in \emph{2010 IEEE
  International Conference on Robotics and Automation}.\hskip 1em plus 0.5em
  minus 0.4em\relax IEEE, 2010, pp. 21--28.

\bibitem{mellinger2011minimum}
D.~Mellinger and V.~Kumar, ``Minimum snap trajectory generation and control for
  quadrotors,'' in \emph{{IEEE} Int. Conf. Robot. Autom. (ICRA)}, 2011, pp.
  2520--2525.

\bibitem{Richter2013ISRR}
C.~Richter, A.~Bry, and N.~Roy, ``Polynomial trajectory planning for aggressive
  quadrotor flight in dense indoor environments,'' 2013.

\bibitem{foehn2021time}
P.~Foehn, A.~Romero, and D.~Scaramuzza, ``Time-optimal planning for quadrotor
  waypoint flight,'' \emph{Science Robotics}, vol.~6, no.~56, p. eabh1221,
  2021.

\bibitem{liu2018search}
S.~Liu, K.~Mohta, N.~Atanasov, and V.~Kumar, ``Search-based motion planning for
  aggressive flight in se (3),'' \emph{IEEE Robotics and Automation Letters},
  vol.~3, no.~3, pp. 2439--2446, 2018.

\bibitem{song2022policy}
Y.~Song and D.~Scaramuzza, ``Policy search for model predictive control with
  application to agile drone flight,'' \emph{IEEE Transactions on Robotics},
  pp. 1--17, 2022.

\bibitem{faessler2017differential}
M.~Faessler, A.~Franchi, and D.~Scaramuzza, ``Differential flatness of
  quadrotor dynamics subject to rotor drag for accurate tracking of high-speed
  trajectories,'' \emph{IEEE Robotics and Automation Letters}, vol.~3, no.~2,
  pp. 620--626, 2017.

\bibitem{adamkiewicz22ral}
M.~Adamkiewicz, T.~Chen, A.~Caccavale, R.~Gardner, P.~Culbertson, J.~Bohg, and
  M.~Schwager, ``Vision-only robot navigation in a neural radiance world,''
  \emph{IEEE Robotics and Automation Letters}, vol.~7, no.~2, pp. 4606--4613,
  2022.

\bibitem{mildenhall2021nerf}
B.~Mildenhall, P.~P. Srinivasan, M.~Tancik, J.~T. Barron, R.~Ramamoorthi, and
  R.~Ng, ``Nerf: Representing scenes as neural radiance fields for view
  synthesis,'' \emph{Communications of the ACM}, vol.~65, no.~1, pp. 99--106,
  2021.

\bibitem{ross2013learning}
S.~Ross, N.~Melik-Barkhudarov, K.~S. Shankar, A.~Wendel, D.~Dey, J.~A. Bagnell,
  and M.~Hebert, ``Learning monocular reactive uav control in cluttered natural
  environments,'' in \emph{2013 IEEE international conference on robotics and
  automation}.\hskip 1em plus 0.5em minus 0.4em\relax IEEE, 2013, pp.
  1765--1772.

\bibitem{sadeghi2016cad2rl}
F.~Sadeghi and S.~Levine, ``Cad2rl: Real single-image flight without a single
  real image,'' \emph{arXiv preprint arXiv:1611.04201}, 2016.

\bibitem{zhang2016learning}
T.~Zhang, G.~Kahn, S.~Levine, and P.~Abbeel, ``Learning deep control policies
  for autonomous aerial vehicles with mpc-guided policy search,'' in \emph{2016
  IEEE international conference on robotics and automation (ICRA)}.\hskip 1em
  plus 0.5em minus 0.4em\relax IEEE, 2016, pp. 528--535.

\bibitem{sampedro18icra}
C.~Sampedro, A.~Rodriguez-Ramos, I.~Gil, L.~Mejias, and P.~Campoy,
  ``Image-based visual servoing controller for multirotor aerial robots using
  deep reinforcement learning,'' in \emph{2018 IEEE/RSJ International
  Conference on Intelligent Robots and Systems (IROS)}, 2018, pp. 979--986.

\bibitem{devo2022ral}
A.~Devo, J.~Mao, G.~Costante, and G.~Loianno, ``Autonomous single-image drone
  exploration with deep reinforcement learning and mixed reality,'' \emph{IEEE
  Robotics and Automation Letters}, vol.~7, no.~2, 2022.

\bibitem{singla21its}
\BIBentryALTinterwordspacing
A.~Singla, S.~Padakandla, and S.~Bhatnagar, ``Memory-based deep reinforcement
  learning for obstacle avoidance in uav with limited environment knowledge,''
  \emph{Trans. Intell. Transport. Syst.}, vol.~22, no.~1, p. 107–118, jan
  2021. [Online]. Available: \url{https://doi.org/10.1109/TITS.2019.2954952}
\BIBentrySTDinterwordspacing

\bibitem{kim22esa}
\BIBentryALTinterwordspacing
M.~Kim, J.~Kim, M.~Jung, and H.~Oh, ``Towards monocular vision-based autonomous
  flight through deep reinforcement learning,'' \emph{Expert Systems with
  Applications}, vol. 198, p. 116742, 2022. [Online]. Available:
  \url{https://www.sciencedirect.com/science/article/pii/S0957417422002111}
\BIBentrySTDinterwordspacing

\bibitem{hinton2006reducing}
G.~E. Hinton and R.~R. Salakhutdinov, ``Reducing the dimensionality of data
  with neural networks,'' \emph{science}, vol. 313, no. 5786, pp. 504--507,
  2006.

\bibitem{yunlong2020flightmare}
Y.~Song, S.~Naji, E.~Kaufmann, A.~Loquercio, and D.~Scaramuzza, ``Flightmare: A
  flexible quadrotor simulator,'' in \emph{Conference on Robot Learning}, 2020.

\bibitem{schulman2017proximal}
J.~Schulman, F.~Wolski, P.~Dhariwal, A.~Radford, and O.~Klimov, ``Proximal
  policy optimization algorithms,'' \emph{arXiv preprint arXiv:1707.06347},
  2017.

\bibitem{zhou2019robust}
B.~Zhou, F.~Gao, L.~Wang, C.~Liu, and S.~Shen, ``Robust and efficient quadrotor
  trajectory generation for fast autonomous flight,'' \emph{IEEE Robotics and
  Automation Letters}, vol.~4, no.~4, pp. 3529--3536, 2019.

\bibitem{florence2020integrated}
P.~Florence, J.~Carter, and R.~Tedrake, ``Integrated perception and control at
  high speed: Evaluating collision avoidance maneuvers without maps,'' in
  \emph{Algorithmic Foundations of Robotics XII}.\hskip 1em plus 0.5em minus
  0.4em\relax Springer, 2020, pp. 304--319.

\bibitem{aFalanga19ral_howfast}
D.~Falanga, S.~Kim, and D.~Scaramuzza, ``How fast is too fast? the role of
  perception latency in high-speed sense and avoid,'' \emph{{IEEE} Robot.
  Autom. Lett.}, vol.~4, no.~2, pp. 1884--1891, Apr. 2019.

\bibitem{foehn2022agilicious}
P.~Foehn, E.~Kaufmann, A.~Romero, R.~Penicka, S.~Sun, L.~Bauersfeld,
  T.~Laengle, G.~Cioffi, Y.~Song, A.~Loquercio \emph{et~al.}, ``Agilicious:
  Open-source and open-hardware agile quadrotor for vision-based flight,''
  \emph{Science Robotics}, vol.~7, no.~67, p. eabl6259, 2022.

\bibitem{penicka2022minimum}
R.~Penicka and D.~Scaramuzza, ``Minimum-time quadrotor waypoint flight in
  cluttered environments,'' \emph{IEEE Robotics and Automation Letters},
  vol.~7, no.~2, pp. 5719--5726, 2022.

\bibitem{Foehn2021science}
\BIBentryALTinterwordspacing
P.~Foehn, A.~Romero, and D.~Scaramuzza, ``Time-optimal planning for quadrotor
  waypoint flight,'' \emph{Science Robotics}, vol.~6, no.~56, 2021. [Online].
  Available: \url{https://robotics.sciencemag.org/content/6/56/eabh1221}
\BIBentrySTDinterwordspacing

\bibitem{Sun2021tro}
S.~Sun, A.~Romero, P.~Foehn, E.~Kaufmann, and D.~Scaramuzza, ``A comparative
  study of nonlinear mpc and differential-flatness-based control for quadrotor
  agile flight,'' \emph{IEEE Transactions on Robotics}, 2022.

\bibitem{hochreiter1997long}
S.~Hochreiter and J.~Schmidhuber, ``Long short-term memory,'' \emph{Neural
  computation}, vol.~9, no.~8, pp. 1735--1780, 1997.

\bibitem{vaswani2017attention}
A.~Vaswani, N.~Shazeer, N.~Parmar, J.~Uszkoreit, L.~Jones, A.~N. Gomez,
  {\L}.~Kaiser, and I.~Polosukhin, ``Attention is all you need,''
  \emph{Advances in neural information processing systems}, vol.~30, 2017.

\end{thebibliography}
